\documentclass[final,3p, times]{elsarticle}  

\usepackage{amssymb}
\usepackage{amsmath}

\begin{document}

\begin{frontmatter}


\title{Towards Explainable Skin Cancer Classification: A Dual-Network Attention Model with Lesion Segmentation and Clinical Metadata Fusion} 


\author{Md Enamul Atiq} 
\author{Shaikh Anowarul Fattah}

\affiliation{organization={Department of Electrical and Computer Engineering, Bangladesh University of Engineering and Technology}, city={Dhaka}, country={Bangladesh}}

\begin{abstract}
Skin cancer is a life-threatening disease where early detection significantly improves patient outcomes. Automated diagnosis from dermoscopic images is challenging due to high intra-class variability and subtle inter-class differences. Many deep learning models operate as "black boxes", limiting clinical trust. In this work, we propose a dual-encoder attention-based framework that leverages both segmented lesions and clinical metadata to enhance skin lesion classification in terms of both accuracy and interpretability. A novel Deep-UNet architecture with Dual Attention Gates (DAG) and Atrous Spatial Pyramid Pooling (ASPP) is first employed to segment lesions. The classification stage uses two DenseNet201 encoders, one on the original image and another on the segmented lesion whose features are fused via multi-head cross-attention. This dual-input design guides the model to focus on salient pathological regions. In addition, a transformer-based module incorporates patient metadata (age, sex, lesion site) into the prediction. We evaluate our approach on the HAM10000 dataset and the ISIC 2018 and 2019 challenges. The proposed method achieves state-of-the-art segmentation performance and significantly improves classification accuracy and average AUC compared to baseline models. To validate our model's reliability, we use Gradient-weighted Class Activation Mapping (Grad-CAM) to generate heatmaps. These visualizations confirm that our model's predictions are based on the lesion area, unlike models that rely on spurious background features. These results demonstrate that integrating precise lesion segmentation and clinical data with attention-based fusion leads to a more accurate and interpretable skin cancer classification model.
\end{abstract}



\begin{keyword}


Lesion, Segmentation, Dermoscopy, CNN, Multihead, Cross-attention, Heatmap, Grad-CAM
\end{keyword}

\end{frontmatter}



\section{Introduction}
\label{sec1}

Skin cancer is one of the most prevalent forms of cancer worldwide, with millions of new diagnoses each year. Early and accurate identification of skin lesions is crucial for effective treatment and improved patient outcomes. Dermoscopic imaging, which offers magnified, polarized views of skin lesions, is widely used by dermatologists to assess potentially cancerous regions. However, interpreting dermoscopic images remains challenging due to subtle visual differences between lesion types, overlapping features across classes, and common artifacts (e.g., hair, bubbles) that can obscure lesion boundaries.

Deep learning, particularly convolutional neural networks (CNNs), has shown strong potential in automating skin lesion analysis. CNN-based models have achieved performance comparable to dermatologists in certain settings. Yet, several issues limit their clinical adoption. Lesions often cover only a small image region, with surrounding skin or artifacts introducing noise. Moreover, the "black-box" nature of many deep learning models makes their decision-making process opaque, which is a significant barrier to establishing clinical trust. Without a clear understanding of which features a model is using, it is difficult to verify its reliability. Most models also ignore useful patient metadata (such as age, sex, or lesion location), which could aid diagnosis.

To address these challenges, our work focuses on two key areas: improving classification accuracy by forcing the model to concentrate on relevant regions, and enhancing model transparency to build trust in its predictions. Lesion segmentation helps mitigate these challenges by isolating the lesion from the background. By providing the model with both the original image and its segmented mask, we can explicitly guide its focus. Furthermore, to verify that our model is indeed focusing on the lesion, we employ explainability techniques like Gradient-weighted Class Activation Mapping (Grad-CAM). Grad-CAM generates visual heatmaps that highlight the image regions most influential in a model's prediction, offering a way to peer inside the "black box" and validate its reasoning.

In this work, we present a unified framework that integrates segmentation and clinical metadata for robust and interpretable multi-class skin cancer classification. Our key contributions are as follows:
\begin{itemize}
    \item We propose a Deep-UNet segmentation network with \emph{Dual Attention Gate (DAG)} modules and an \emph{Atrous Spatial Pyramid Pooling (ASPP)} block. It achieves accurate lesion localization using supervised segmentation methods.
    \item We develop a dual-encoder classification architecture using two DenseNet201 encoders, one for the original image and one for the segmented lesion. A \emph{multi-head cross-attention} module fuses these features, enabling the model to attend to both contextual and localized information, thereby improving accuracy.
    \item We introduce a clinical metadata transformer that encodes patient-specific information, which is fused with image features to enhance prediction reliability.
    \item We validate our model's decision-making process using \textbf{Grad-CAM}. We provide visual evidence that our dual-input approach forces the model to focus on the segmented lesion, unlike standard models that may rely on irrelevant background features. This significantly enhances the model's explainability and trustworthiness.
    \item We evaluate our comprehensive framework on \textbf{HAM10000}, \textbf{ISIC 2018 Challenge Task 1}, and \textbf{ISIC 2019}, achieving high performance in both segmentation and multi-class classification, along with ablation studies validating each component’s contribution.
\end{itemize}

\section{Literature Review}
\label{literature}

The automated analysis of skin lesions has gained significant attention in recent years, driven by advances in deep learning and the availability of large-scale dermoscopic datasets. This section reviews the current state-of-the-art in skin cancer classification, lesion segmentation, and multimodal approaches that combine imaging data with clinical metadata.

\subsection{Lesion Segmentation in Dermoscopy}
\label{subsec:segmentation}

Accurate lesion segmentation is crucial for effective skin cancer classification, as it helps isolate the region of interest from background skin, hair, and other artifacts. Traditional segmentation methods relied on threshold-based approaches and region growing techniques, which often failed to handle the complex variations in dermoscopic images.

One of the foundational models is U-Net, proposed by \citet{ronneberger2015u}, which employs a symmetric encoder-decoder architecture with skip connections to preserve spatial information during segmentation. This architecture has become a cornerstone in biomedical image segmentation due to its robustness and adaptability. Building on this foundation, \citet{zhou2018unet++} introduced U-Net++, which incorporates nested and dense skip connections to reduce the semantic gap between encoder and decoder feature maps. This design improves performance on datasets with complex boundaries and variable lesion scales. DeepLabV3+, developed by \citet{Chen_2018_ECCV}, enhances segmentation accuracy through atrous spatial pyramid pooling and an encoder-decoder framework. Its ability to capture multi-scale contextual information has made it widely applicable in both natural and medical imaging domains.

Recent studies have explored hybrid and transformer-based models to further improve segmentation accuracy. \citet{hu2024skinsam} introduced SkinSAM, which adapts the Segment Anything Model (SAM) with a ViT backbone, fine-tuned specifically for the HAM10000 dataset. \citet{roy2025melseg} developed melSeg, an adaptation of SAM using a feature pyramid network to capture multi-scale lesion features, exhibiting strong performance on HAM10000.

\citet{naveed2024ra} presented RA-Net, a region-aware attention network that integrates guided loss within an encoder-decoder structure, achieving consistent performance across ISIC and PH2 datasets. Similarly, \citet{lan2024brau} proposed BRAU-Net++, a CNN-Transformer hybrid model employing bi-level routing attention and channel-spatial attention mechanisms. 

\citet{ahmed2025precision, ahmed2025hybrid} proposed two dual-encoder models—DuaSkinSeg, which combines MobileNetV2 and ViT-CNN, and a hybrid model using Squeeze-and-Excitation (SE) attention blocks—demonstrating strong performance across various ISIC datasets. Another approach by \citet{alhudhaif2022novel} introduced a Multipath Fusion Model (MF-Model), which utilizes three parallel encoder paths with different convolutional strategies to effectively capture multi-scale features. Their model also introduced a novel "fusion loss" function, combining Binary Cross-Entropy and Jaccard loss to improve performance on the ISIC 2016, 2017, and 2018 datasets.

These works collectively demonstrate the evolution from classical encoder-decoder architectures to attention-augmented and transformer-based networks, contributing to improved segmentation precision, robustness, and computational efficiency.

\subsection{Deep Learning for Skin Cancer Classification}
\label{subsec:classification}

Early approaches to automated skin cancer diagnosis relied on handcrafted features and traditional machine learning classifiers. However, the emergence of deep convolutional neural networks (CNNs) has revolutionized this field. 

The study by \citet{lan2022fixcaps} introduced FixCaps, a capsule network with enhanced attention and large-kernel convolution, improving classification accuracy while reducing computational complexity. \citet{yang2023diffmic} proposed DiffMIC, a diffusion-based model employing dual-granularity conditional guidance to robustly capture semantic representation and eliminate noise from medical images. \citet{saha2024yolov8} employed a YOLOv8-based deep learning approach for real-time skin lesion classification using the HAM10000 dataset. This method demonstrated the feasibility of using object detection frameworks for skin lesion classification tasks, providing fast and accurate predictions.

\citet{maqsood2023multiclass} developed a deep learning-based framework for skin lesion localization and classification. Their approach employed a fusion of deep features extracted from multiple pre-trained CNN models, showing strong performance on various skin lesion datasets. \citet{patricio2024towards} presented a concept-based interpretability model using vision-language methods, providing a human-understandable explanation for skin lesion diagnosis. This approach leverages inherently interpretable models that align well with clinical decision-making processes. \citet{xin2022improved} proposed an improved transformer network, SkinTrans, for skin cancer classification. The model utilized multi-scale patch embedding and contrastive learning to enhance classification accuracy, demonstrating the potential of transformers in medical image analysis. 

In addition to accuracy, model efficiency is crucial for deployment in smart healthcare and mobile applications. Addressing this, \citet{hoang2022multiclass} proposed a lightweight framework that fused deep features from MobileNetV2 and SqueezeNet. By incorporating an attention mechanism and using Bayesian optimization for hyperparameter tuning, their model achieved high performance with significantly fewer parameters. Similarly, \citet{lungu2023skindistilvit} developed SkinDistilViT, a lightweight vision transformer that retains high performance through knowledge distillation. These studies highlight a growing trend towards creating resource-efficient models suitable for real-world clinical environments.

\subsection{Multimodal Approaches and Clinical Data Integration}
\label{subsec:multimodal}

Combining segmentation and classification has the potential to improve diagnostic performance by ensuring the classifier focuses on the lesion region. Besides incorporating clinical metadata can significantly enhance diagnostic accuracy. Patient demographics (age, sex), lesion location, and medical history provide valuable contextual information that dermatologists naturally consider during diagnosis.

\citet{anand2023fusion} integrated U-Net and CNN models for segmentation and classification of skin lesions, showcasing the benefits of fusion models in enhancing classification performance. Their fusion model leveraged the segmentation capabilities of U-Net with the classification power of CNN, offering a robust solution for medical image analysis. \citet{mustafa2025deep} proposed a hybrid approach combining ResUNet++ for segmentation and a modified AlexNet-Random Forest model for classification. The framework enhanced segmentation and classification accuracy, outpacing previous models, and showed its applicability in improving skin cancer diagnosis.

\citet{ahammed2022machine} proposed a machine learning-based approach combining image segmentation and classification techniques to predict various skin diseases accurately. Their method included preprocessing steps like digital hair removal, segmentation with the Grabcut technique, and classification using machine learning algorithms. \citet{shinde2024dermsegnet} introduced DermSegNet, a hybrid model combining adaptive GrabCut segmentation with EfficientNetB3 for multi-class dermatological lesion diagnosis, establishing a new benchmark in dermatological image classification.

\section{Materials}\label{materials}

\subsection{Datasets}\label{sec3.1}
We evaluate our classification approach on two publicly available datasets of dermoscopic images: HAM10000 and ISIC 2019.

\begin{itemize}
    \item\textbf{HAM10000 (Human Against Machine with 10000 Training Images)} \cite{tschandl2018ham10000} is a dataset of 10,015 dermoscopic images, which has been widely used for training skin lesion classifiers. The images in HAM10000 are categorized into 7 diagnostic classes: melanocytic nevus (NV), melanoma (MEL), basal cell carcinoma (BCC), actinic keratosis (AKIEC), benign keratosis (BKL, including solar lentigo/seborrheic keratosis), dermatofibroma (DF), and vascular lesion (VASC). The class distribution is highly imbalanced; for example, about 70\% of the images are nevi, while fewer than 2\% are dermatofibromas. HAM10000 also provides meta-information for each lesion, including the patient’s age, sex, and the general anatomic site of the lesion (e.g., torso, lower extremity, etc.). We utilize these metadata in our model's clinical data branch.

    \item\textbf{ISIC 2019} is the challenge dataset from $2019$ which significantly expands the number of images and classes \cite{andrewmvd2019isic}. The ISIC 2019 training set contains 25,331 dermoscopic images categorized into 8 classes. These include the 7 classes from HAM10000 (NV, MEL, BCC, AKIEC, BKL, DF, VASC) and one additional class: squamous cell carcinoma (SCC). The dataset also provided patient metadata similar to HAM10000.
\end{itemize}

Besides we evaluate our proposed segmentation model on a dataset-

\begin{itemize}
    \item\textbf{ISIC 2018 Challenge Task 1} is an international challenge dataset from the $2018$ ISIC challenge \cite{tschandl2018isic2018}. The ISIC 2018 segmentation dataset contains 2,594 dermoscopic images with corresponding manually annotated lesion masks. The images are similar in content to HAM10000 (and indeed, many HAM10000 images are included in this challenge) but with pixel-level labels. We use this set to train and evaluate our supervised segmentation model.
\end{itemize}

All images across these datasets were RGB dermoscopic photographs. The ground truth for segmentation (ISIC 2018 Task 1) is a binary mask per image, and for classification (HAM10000/ISIC2019) is a single label per image. The metadata available (age, sex, location) were used for all images.

\section{Preprocessing}\label{Preprocessing}

To ensure a consistent input resolution and improve generalization. All images and corresponding masks were resized to $224 \times 224$ pixels. Image normalization was performed using ImageNet mean [0.485, 0.456, 0.406] and standard deviation[0.229, 0.224, 0.225].

\subsection{Segmentation}\label{Segmentation}

Skin lesion segmentation is a critical step in automated skin cancer diagnosis, enabling precise lesion localization and facilitating accurate classification.

Convolutional Neural Networks (CNNs), especially encoder-decoder architectures such as U-Net, have achieved significant success in biomedical segmentation tasks. However, deeper networks often require substantial computational resources, limiting their applicability in clinical environments with constrained hardware. Additionally, lesion segmentation remains challenging due to significant variability in lesion appearance, size, and boundaries.

This work proposes an optimized Deep-UNet architecture designed to balance segmentation accuracy with computational efficiency. The architecture enhances the traditional U-Net by integrating Depthwise Attention Gated (DAG) blocks for refined skip connections and an Atrous Spatial Pyramid Pooling (ASPP) module for multi-scale feature extraction. The model is trained and evaluated on a publicly available datasets, ISIC2018 Challenge Task 1, demonstrating superior performance over state-of-the-art segmentation networks in terms of Dice coefficient and memory footprint.

The contributions of this paper are threefold: (1) a novel Deep-UNet architecture incorporating DAG and ASPP modules; (2) comprehensive evaluation on benchmark dermatological datasets; and (3) demonstration of improved segmentation accuracy with small model size, making it suitable for real-world deployment.

\subsubsection{Deep-UNet Architecture}
\begin{figure}[h]
    \centering
    \includegraphics[width=0.98\textwidth]{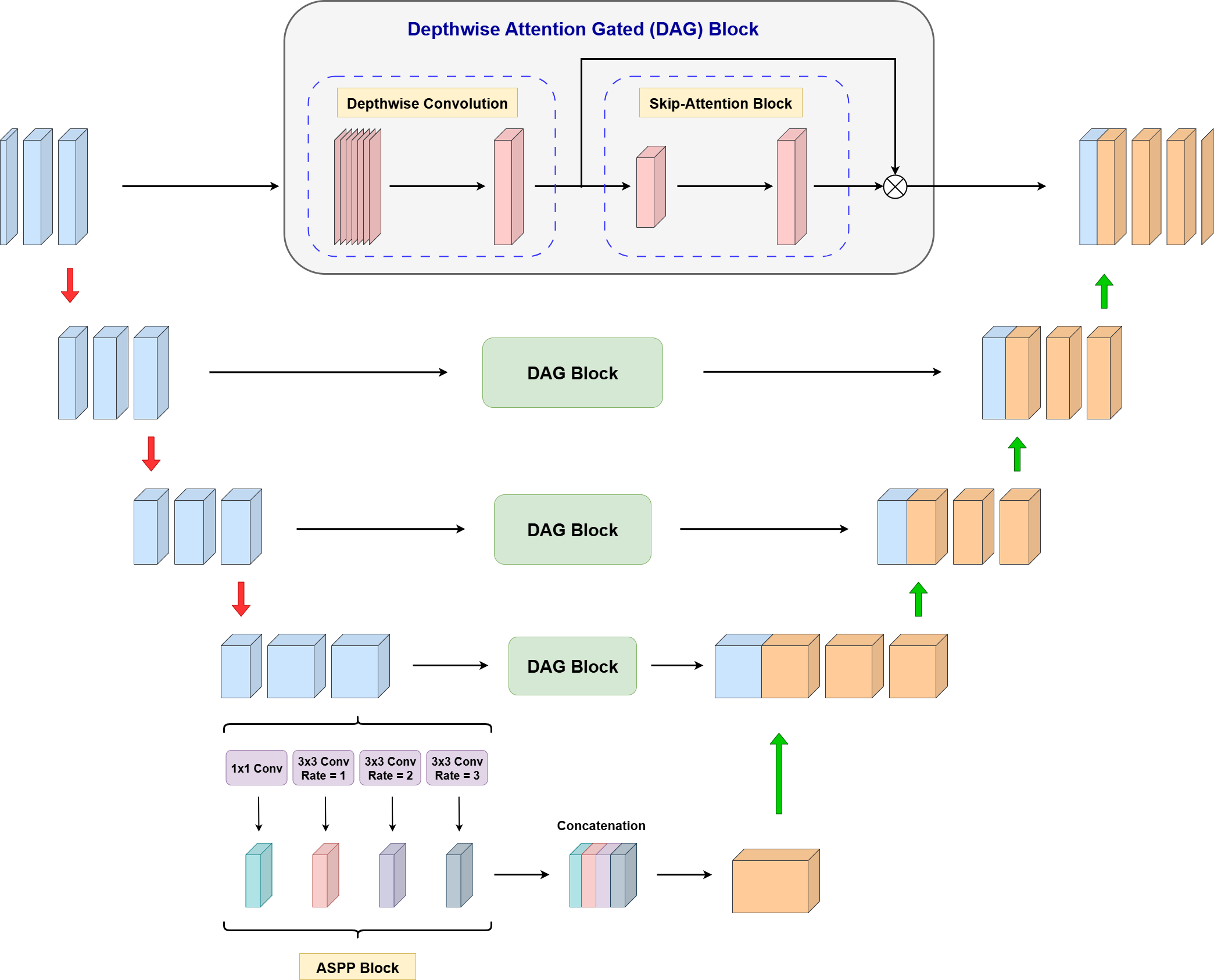}
    \caption{Overview of the proposed Deep-UNet architecture for skin lesion segmentation. The model uses an EfficientNet-B3 encoder, an ASPP bottleneck for multi-scale feature extraction, and a decoder with Depthwise Attention Gated (DAG) skip connections and deep supervision outputs.}
    \label{fig:deepunet_arch}
\end{figure}

The proposed \textbf{Deep-UNet} model is an encoder–decoder network built upon the U-Net framework, with several enhancements to improve segmentation performance (Fig.~\ref{fig:deepunet_arch}). The encoder uses EfficientNet-B3 as a backbone, initialized with ImageNet weights for effective feature extraction. This encoder produces a hierarchy of feature maps at multiple scales. At the encoder–decoder bottleneck, we integrate an \textit{Atrous Spatial Pyramid Pooling} (ASPP) module. The ASPP consists of parallel atrous (dilated) convolutions with different dilation rates (e.g., 1, 2, and 3) and a $1\times1$ convolution, whose outputs are concatenated and fused. This helps capture multi-scale context and enhances the detection of lesions of varying sizes.

On the decoder side, Deep-UNet introduces novel \textit{Depthwise Attention Gated} (DAG) blocks in each skip connection. Each DAG block comprises two main components: a Depthwise Separable Convolution (DSC) for efficient, fine-grained feature processing, and a Skip Attention mechanism. The skip attention is implemented as a small convolutional sub-network (with sigmoid activation) that generates an attention mask over the encoder’s feature map. This mask highlights salient lesion regions and suppresses background before the skip connection is merged. The refined encoder features are then concatenated with the upsampled decoder features. Each decoder stage performs upsampling via a transpose convolution, followed by a \textit{double convolution} block (two successive $3\times3$ conv-BN-ReLU layers) to further refine the fused features. Notably, depthwise separable convolutions are used in place of standard convolutions in these blocks to reduce the parameter count and computational cost. The decoder progressively recovers spatial resolution, and the final $1\times1$ convolution layer produces the output segmentation mask.

In addition, we employ \textit{deep supervision} to facilitate training. Auxiliary output layers are attached to two intermediate decoder levels, each generating an upsampled prediction of the lesion mask. These auxiliary outputs are only used during training to provide additional guidance. By enforcing segmentation predictions at multiple scales, the network’s deeper layers learn more discriminative features early on. During inference, only the final output from the last decoder layer is used for the segmentation result.

\subsubsection{Training Configuration}
The model was implemented in PyTorch and trained end-to-end on an NVIDIA Tesla P100 GPU (16\,GB VRAM). The training configuration included a batch size of 32, with the model trained over 12 epochs. The following parameters and techniques were applied to ensure optimal training and convergence:

\textbf{Loss Function:} We optimized a hybrid loss function combining Dice loss and binary cross-entropy (BCE) loss. For a predicted mask $P$ and ground-truth mask $G$, the Dice loss is defined as-
\begin{equation}
\text{Dice Loss} = 1 - \frac{2 \sum_{i=1}^{N} y_i p_i + \epsilon}{\sum_{i=1}^{N} y_i + \sum_{i=1}^{N} p_i + \epsilon}
\end{equation}
and the BCE loss is-
\begin{equation}
\text{BCE} = -\frac{1}{N}\sum_{i=1}^{N}[\,y_i \log p_i + (1 - y_i)\log(1 - p_i)\,]
\end{equation}
where $p_i$ is the predicted probability for pixel $i$ belonging to the lesion class. The total loss for the final output is-
\begin{equation}
\mathcal{L}_{\text{main}} = \text{BCE} + \text{Dice Loss}
\end{equation}
For the deep supervision outputs, we compute the same loss at each auxiliary output. During training, the losses from the two auxiliary outputs are added to the main loss with weighting factors of 0.2 and 0.1, respectively, to form the overall objective.
\begin{align}
\mathcal{L}_{\text{total}} &= \left( \text{BCE}_{\text{final}} + \text{Dice Loss}_{\text{final}} \right) + 0.2 \left( \text{BCE}_{\text{aux1}} + \text{Dice Loss}_{\text{aux1}} \right) + 0.1 \left( \text{BCE}_{\text{aux2}} + \text{Dice Loss}_{\text{aux2}} \right)
\end{align}

This weighted auxiliary loss strategy provides additional gradient signals to early decoder layers while keeping the final prediction loss dominant.

\textbf{Optimizer:} We used the Adam optimizer with an initial learning rate of $1\times10^{-4}$. A learning rate scheduler reduced the learning rate by half every 4 epochs to ensure stable convergence as training progressed. 

The training/validation split was 80/20\% for each dataset, and the model with the best validation Dice score was selected for final evaluation.

\subsubsection{Experiments and Results}
We evaluated the segmentation performance of Deep-UNet on the test sets of HAM10000 and ISIC 2018. Two standard metrics were used: the Dice coefficient and the Intersection over Union (IoU).

Table~\ref{tab:results} compares our proposed Deep-UNet with other segmentation models, including the original U-Net, the improved U-Net++ architecture, and DeepLabV3+. For fairness, all models in this comparison use an EfficientNet-B3 encoder backbone (except Deep-UNet, which inherently uses EfficientNet-B3). We report the mean Dice (mDice) and mean IoU (mIoU) achieved on each dataset, as well as the model size (memory footprint) to indicate complexity. As shown in the table, Deep-UNet achieves an mDice of 91.17\% and mIoU of 84.35\% on the ISIC 2018 dataset. These results are on par with the best-performing baseline (U-Net++ with EfficientNet-B3, 91.54\% Dice) and substantially better than the original U-Net and DeepLabV3+. Notably, Deep-UNet attains this accuracy with a significantly smaller model size (about 204~MB) compared to U-Net++ (273~MB) and DeepLabV3+ (211~MB). In particular, while U-Net++ slightly exceeds Deep-UNet in Dice by 0.37\%, it uses roughly 34\% more memory. Deep-UNet’s efficient design thus offers an excellent trade-off between segmentation accuracy and model complexity. The strong performance across both datasets demonstrates the effectiveness of our architectural contributions, the EfficientNet-B3 encoder, ASPP bottleneck, and attention-enhanced decoder with deep supervision in improving skin lesion segmentation. A figure illustrating the segmentation process for the HAM10000 and ISIC 2019 dataset is shown below:

\begin{figure}[h]
    \centering
    \includegraphics[width=1\textwidth]{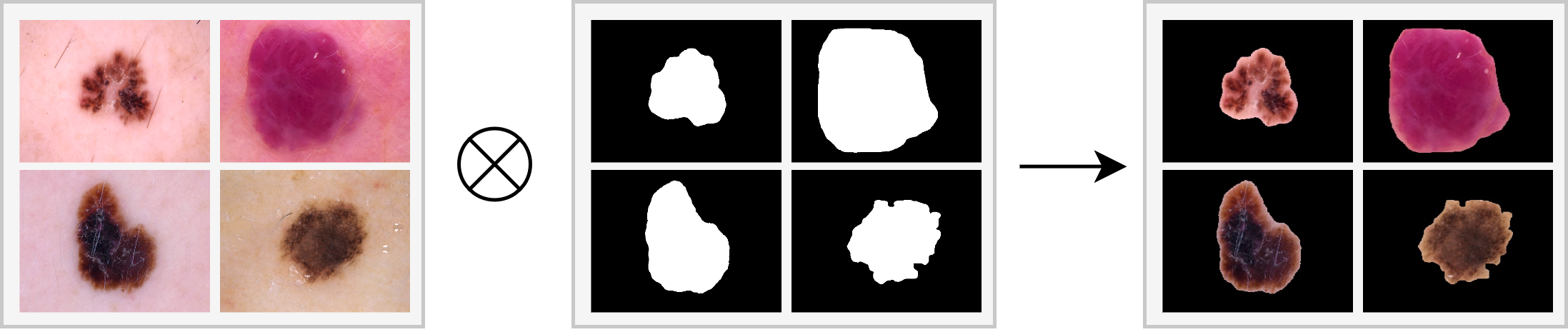}
    \caption{Segmentation Process: Original image multiplied by binary mask to create the segmented image.}
    \label{fig:segmentation_ham10000}
\end{figure}

\begin{table}[h]
\caption{Segmentation performance on ISIC 2018 dataset. All models use EfficientNet-B3 as the encoder backbone.}
\label{tab:results}
\centering
\begin{tabular}{lccc}
\hline
\textbf{Method} & \textbf{mDice} & \textbf{mIoU} & \textbf{Size (MB)} \\
\hline
U-Net          & 90.47\% & 82.71\% & 200.44 \\
U-Net++        & 91.54\% & 84.44\% & 273.32 \\
DeepLabV3+     & 90.11\% & 82.14\% & 211.31 \\
\textbf{Deep-UNet (Ours)} & 91.17\% & 84.35\% & 204.26 \\
\hline
\end{tabular}
\end{table}

\subsection{Dataset Balancing and Augmentation}
Both the HAM10000 and ISIC 2019 datasets exhibited class imbalance, with certain lesion types being underrepresented. To mitigate this issue, dataset balancing and augmentation techniques were employed, ensuring a more uniform class distribution and improving model generalization.

\begin{figure}[h]
    \centering
    \begin{minipage}{0.48\textwidth}
        \centering
        \includegraphics[width=\textwidth]{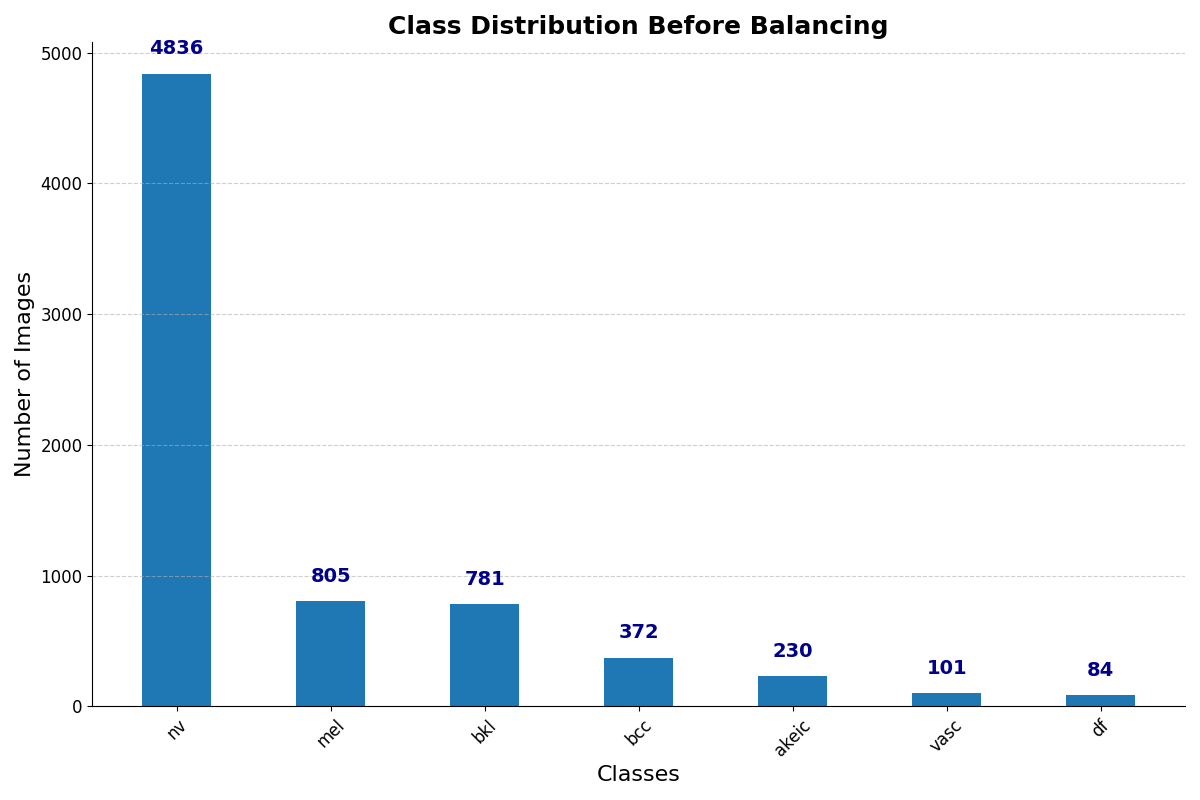}
    \end{minipage}
    \hfill
    \begin{minipage}{0.48\textwidth}
        \centering
        \includegraphics[width=\textwidth]{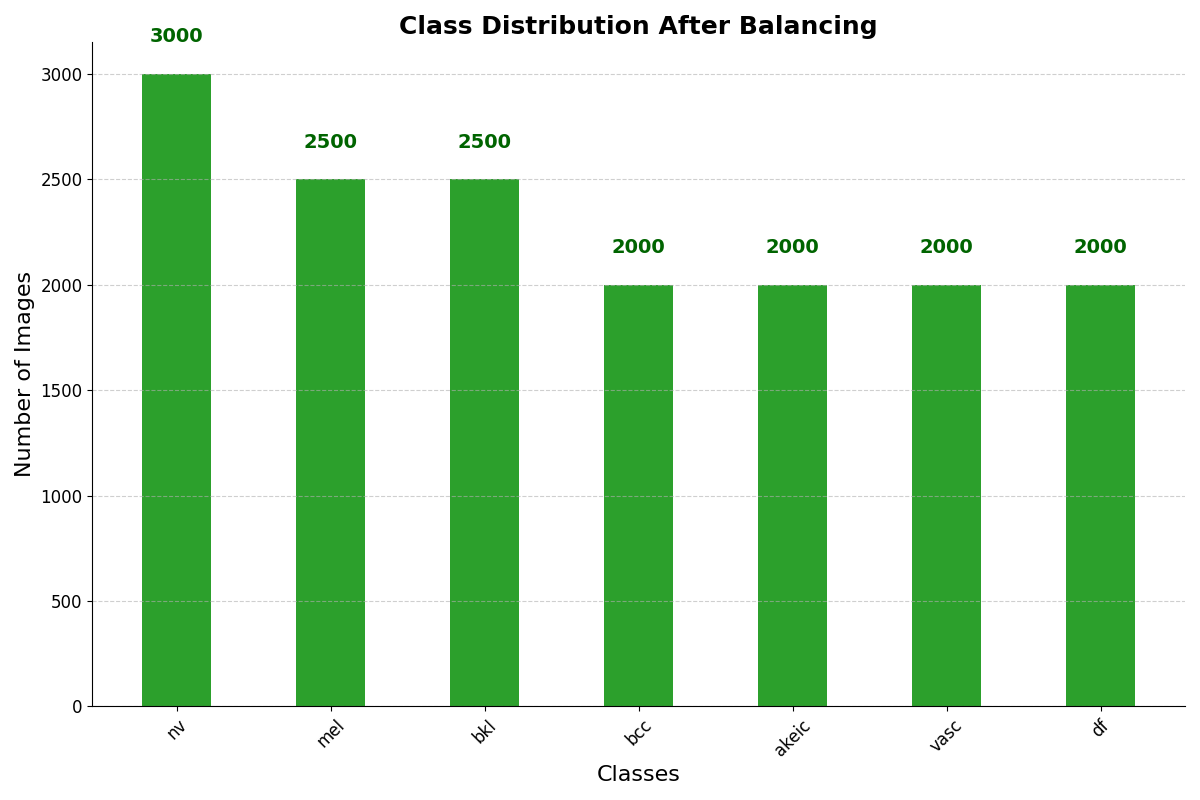}
    \end{minipage}
    \caption{Class distribution in the training dataset before (left) and after (right) balancing for the HAM10000 dataset.}
    \label{fig:before_after_balancing_ham}
\end{figure}

\begin{figure}[h]
    \centering
    \begin{minipage}{0.48\textwidth}
        \centering
        \includegraphics[width=\textwidth]{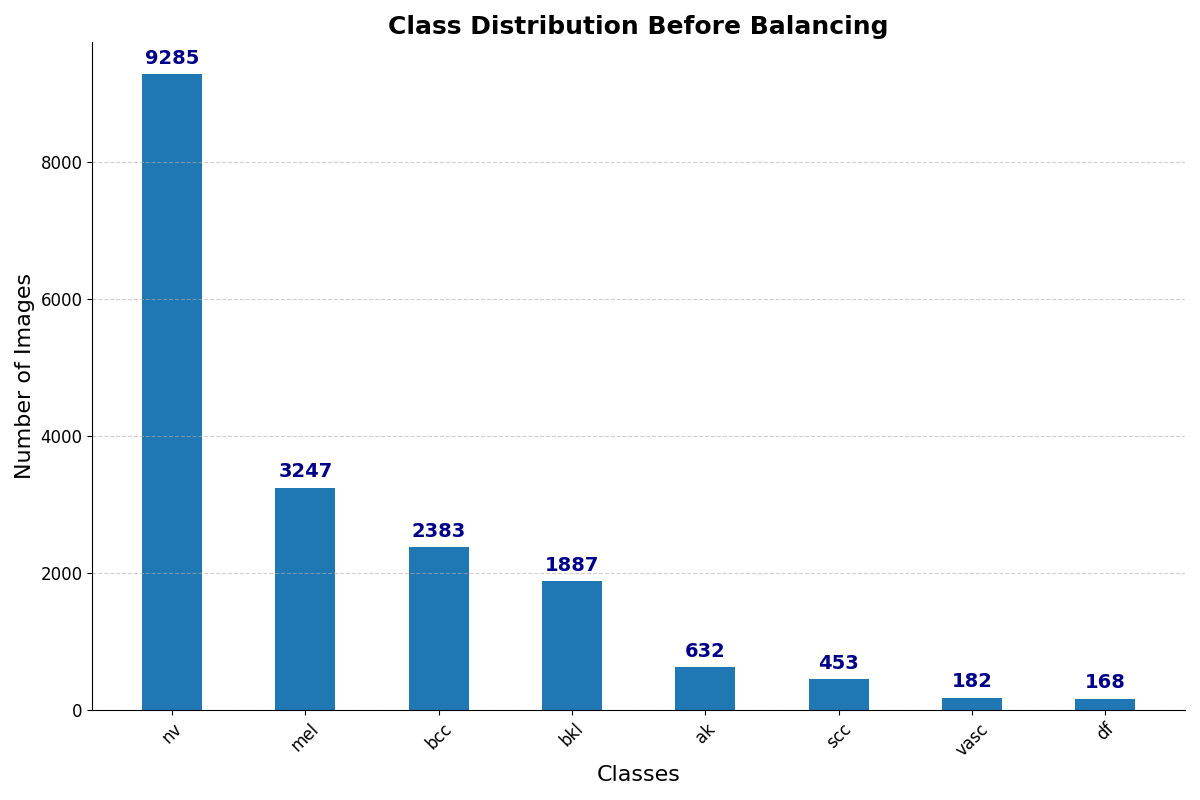}
    \end{minipage}
    \hfill
    \begin{minipage}{0.48\textwidth}
        \centering
        \includegraphics[width=\textwidth]{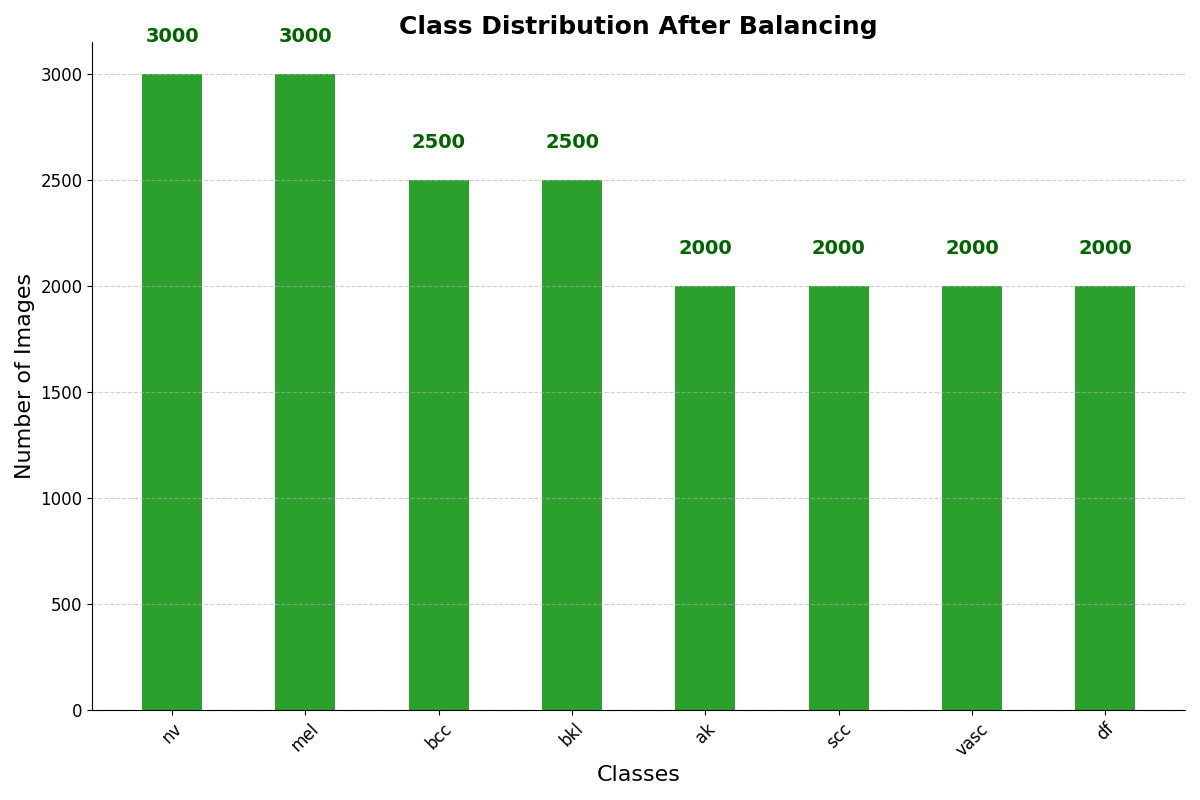}
    \end{minipage}
    \caption{Class distribution in the training dataset before (left) and after (right) balancing for the ISIC 2019 dataset.}
    \label{fig:before_after_balancing_isic}
\end{figure}

Figure~\ref{fig:before_after_balancing_ham} shows the original class distribution of HAM10000 dataset (left) where the \texttt{nv} class dominates the dataset with significantly more images than the other classes. The balanced dataset distribution (right) illustrates the effect of our applied balancing approach. Similarly, the class distribution in the ISIC 2019 dataset was imbalanced. The original class distribution (before balancing) is shown in Figure~\ref{fig:before_after_balancing_isic} (left). After applying the balancing strategy, the dataset distribution becomes more uniform, as shown in the figure on the right.

The primary objective of this balancing strategy is to create a more even distribution of class samples without distorting the natural dataset characteristics. Unlike traditional oversampling, which duplicates images, this method:
\begin{itemize}
    \item Ensures that majority classes retain their presence while reducing dominance.
    \item Expands smaller classes using augmentation rather than simple duplication, maintaining diversity.
\end{itemize}

Strictly equalizing class distributions can introduce biases, as majority classes contain diverse images, while minority classes may require excessive duplication, leading to overfitting. Instead, a controlled balancing approach was used to enhance model generalization while preserving class diversity. This approach ensures a well-distributed dataset that enhances model learning without introducing artificial bias.

Data augmentation was then employed to enhance diversity in the training set through various transformations:
\begin{itemize}
    \item \textbf{Random Horizontal Flip:} Each image had a 50\% probability of being flipped horizontally, introducing variability in image orientation.
    \item \textbf{Random Rotation:} Images were rotated within a range of ±15 degrees to simulate different viewing angles and improve robustness to rotational variance.
    \item \textbf{Normalization:} All images were normalized using mean values of [0.485, 0.456, 0.406] and standard deviations of [0.229, 0.224, 0.225], aligning the data distribution with that of ImageNet-pretrained models.
\end{itemize}

These augmentations were applied during the training phase (in conjunction with oversampling), while only resizing and normalization were used for the test set to ensure consistent evaluation.

\section{Methodology}

\subsection{Overview}
The proposed architecture, termed \textbf{Transformer Dual-Branch Network (TDBN)}, integrates multimodal data for robust skin lesion classification. The model employs two DenseNet201 encoders, one for original dermoscopic images and another for segmented lesion images to capture complementary visual representations. A multi-head cross-attention mechanism facilitates inter-branch feature interaction, guided by the original image as query and value, and the segmented image as key. Additionally, clinical metadata is encoded via a transformer-based tabular embedding network, projected into a shared feature space with visual features. The fused representation is subsequently classified through a deep fully connected classifier.

\subsection{Dual Visual Encoders}
The model employs two parallel visual encoders based on DenseNet201 backbones pretrained on ImageNet. Each encoder outputs dense spatial features from the final convolutional block, with the classification head replaced by an identity layer to preserve full feature maps.

\textbf{Original Image Encoder:} The encoder for unsegmented dermoscopic images captures the global and contextual visual characteristics of the skin region. Given an input of size $(B, 3, 224, 224)$, the DenseNet201 produces a feature tensor of shape $(B, 1920, 7, 7)$, which is then reshaped into a sequence of 49 tokens of dimension 1920, preserving the spatial context for the attention module.

\textbf{Segmented Image Encoder:} The second encoder operates on lesion-segmented inputs, emphasizing lesion morphology and texture while removing background noise. It follows the same configuration as the original encoder, yielding feature tokens of identical dimensionality $(49 \times 1920)$ for cross-branch attention computation.

\subsection{Multi-Head Cross-Attention Mechanism}
To facilitate contextual alignment between global and lesion-specific features, we design a \textbf{Multi-Head Cross-Attention (MHCA)} module, which allows one visual stream to attend to the other. Specifically, the feature sequence from the original image branch serves as the \textit{query} ($Q$) and \textit{value} ($V$), while the segmented image features serve as the \textit{key} ($K$). This enables the model to identify which regions in the segmented image most effectively support discriminative cues in the original representation.

Formally, each attention head computes:
\begin{equation}
\text{Attention}(Q,K,V) = \text{softmax}\!\left(\frac{QK^T}{\sqrt{d_k}}\right)V,
\end{equation}
where $d_k$ is the dimensionality of the key vectors. The outputs from multiple heads are concatenated and processed through a feed-forward network with residual connections and layer normalization:
\begin{equation}
\hat{Z} = \text{LayerNorm}\!\left(Z + \text{FFN}(Z)\right),
\end{equation}
ensuring stable learning and preserving high-level consistency across feature dimensions. Finally, a global average pooling is applied over the sequence to obtain a unified 1920-dimensional vector.

\subsection{Tabular Embedding Network}
Clinical metadata, including patient age, sex, and anatomical lesion site, is processed through a dedicated \textbf{Tabular Embedding Network} designed to encode non-visual clinical features into a latent representation compatible with the visual feature space.

The metadata vector is passed through a two-layer fully connected neural network with ReLU activation and batch normalization, enabling non-linear feature transformation and improved generalization:
\begin{equation}
E_t = \text{BN}\!\left(\text{ReLU}\!\left(W_2 \, \text{BN}\!\left(\text{ReLU}(W_1 x_t)\right)\right)\right),
\end{equation}
where $x_t$ denotes the input tabular vector and $W_1, W_2$ are learnable weight matrices. This network captures interactions among demographic and anatomical attributes, yielding a compact and informative tabular embedding.

To ensure robustness to missing metadata, a binary availability mask $\alpha_t \in \{0,1\}$ is applied such that unavailable features are replaced with a zero vector:
\begin{equation}
\hat{E}_t = \alpha_t \cdot E_t.
\end{equation}
This mechanism maintains training stability and enables the network to adaptively rely on visual cues when metadata is absent.

\subsection{Feature Projection and Fusion}
Before fusion, both visual and tabular representations are independently refined by projection layers composed of fully connected, batch normalization, ReLU activation, and dropout operations. This step harmonizes scale and improves modality-specific feature quality.

The final fusion is achieved through \textbf{additive integration}:
\begin{equation}
F = F_{\text{vision}} + \alpha \cdot F_{\text{tabular}},
\end{equation}
where $\alpha$ is a binary availability mask ensuring robustness to missing metadata (set to zero when metadata is unavailable). This additive strategy offers improved stability compared to concatenation-based fusion, as it preserves feature dimensionality and prevents dominance of any single modality.

\subsection{Classification Head}
The fused representation $F \in \mathbb{R}^{1920}$ is passed through a hierarchical classifier composed of dense layers of sizes 512 and 128, each followed by batch normalization, ReLU activation, and dropout for regularization. The final layer outputs class logits, which are normalized using the softmax function to yield lesion category probabilities. This design ensures a deep, regularized decision boundary capable of handling high-dimensional fused features effectively.

\begin{figure}[h]
    \centering
    \includegraphics[width=1\textwidth]{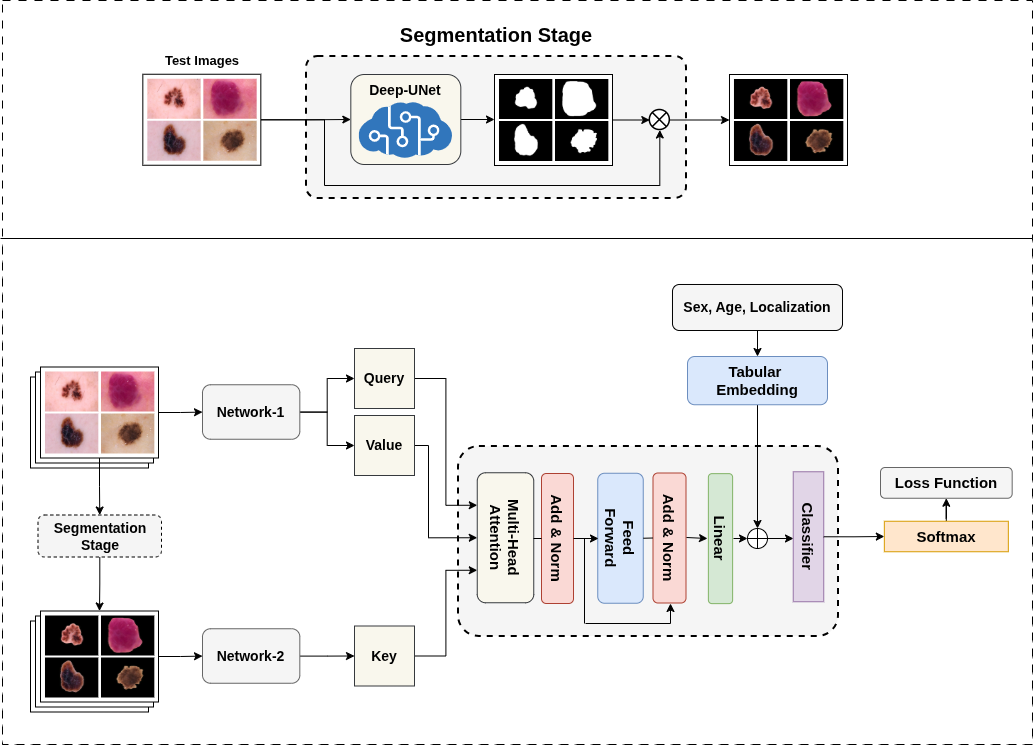}
    \caption{Architecture of the proposed Transformer Dual-Branch Network (TDBN), integrating dual DenseNet encoders, multi-head cross-attention, transformer-based metadata embedding, and additive multimodal fusion.}
    \label{fig:model_architecture}
\end{figure}

\section{Experiment and Results}

\subsection{Training}
The model training was conducted on a Kaggle P100 GPU, which provided accelerated computation. The training configuration included a batch size of 32, with the model trained over 10 epochs. The following parameters and techniques were applied to ensure optimal training and convergence:

\textbf{Loss Function:} Cross-entropy loss was used as the objective function for classification, which is standard for multi-class problems.

\textbf{Optimizer:} The Adam (Adaptive Moment Estimation) Optimizer was used with an initial learning rate of 0.0001, a weight decay of 0.0001, and default values for the beta parameters ($\beta_1=0.9$, $\beta_2=0.999$). Adam optimizes learning rates for each parameter using first and second moment estimates, improving stability and convergence.

\textbf{Learning Rate Scheduler:} A learning rate scheduler was implemented to halve the learning rate every 3 epochs, helping the model converge more smoothly by reducing the learning rate gradually over time.

The cross-attention mechanism introduced an additional computational step during training, requiring pairwise computations between query, key, and value vectors for each input batch. Despite this, convergence was achieved within the same number of epochs as prior methods, demonstrating the efficiency of the implementation.
Each training epoch consisted of multiple steps, including a forward pass through the encoders, multi head cross-attention module, and transformer encoder for clinical metadata. Loss computation, backpropagation, and optimizer updates followed, ensuring adaptive learning of the model parameters. Regular evaluations of accuracy and loss on the validation set provided insights into the model's progression.

Training was completed in a total of 10 epochs, with each epoch’s performance monitored in terms of loss and accuracy metrics to track the model’s learning progression.

\subsection{Results}

To evaluate the effectiveness of the proposed model, we conducted comparative experiments using the HAM10000 and ISIC 2019 datasets. Table~\ref{tab:comparison_models} presents the test accuracy of the proposed model in comparison with several widely-used baseline architectures, including ResNet50, DenseNet201, MobileNetV2, EfficientNetB3, Xception, and ViT.

\begin{table}[t]
\centering
\caption{Comparison of Test Accuracy with Existing Models on HAM10000 and ISIC 2019 Datasets}
\label{tab:comparison_models}
\renewcommand{\arraystretch}{1.2}
\setlength{\tabcolsep}{10pt}
\begin{tabular}{|l|c|c|}
\hline
\textbf{Model} & \textbf{HAM10000 (\%)} & \textbf{ISIC 2019 (\%)} \\
\hline
ResNet50 & 87.12 & 81.45 \\
DenseNet201 & 88.88 & 84.17 \\
Xception & 86.53 & 79.20 \\
MobileNetV2 & 86.78 & 77.86 \\
EfficientNetB3 & 87.98 & 80.68 \\
ViT & 87.98 & 80.99 \\
\hline
\textbf{Proposed Model} & \textbf{93.47} & \textbf{88.56} \\
\hline
\end{tabular}
\end{table}

The proposed model outperforms all baseline models, achieving a test accuracy of 93.47\% on HAM10000 and 88.56\% on ISIC 2019. In addition to accuracy, the model demonstrates strong performance in other metrics such as AUC (0.9900 and 0.9871 respectively) and Top-2 Accuracy (98.40\% and 96.47\% respectively). These results highlight the model's robust ability to generalize across both datasets.

Fig.~\ref{fig:confusion-matrix} shows the confusion matrices for the HAM10000 and ISIC 2019 datasets, respectively. These matrices provide further insight into the model's class-wise performance and indicate balanced sensitivity across most lesion classes.

\subsection{Ablation Study}
To understand the contribution of each component in the proposed model, we conducted an ablation study. The following configurations were tested:

\begin{itemize}
    \item \textbf{DenseNet201 on Original Images Only:} A DenseNet201 model trained on the original images, excluding the segmented images and metadata.
    \item \textbf{DenseNet201 on Segmented Images Only:} A DenseNet201 model trained on the segmented images alone, excluding the original images and metadata.
    \item \textbf{Proposed Dual-Network Model without Metadata:} The full dual-network architecture, excluding the clinical metadata (sex, age, localization).
    \item \textbf{Proposed Dual-Network Model with Metadata:} The complete model, including both visual features from original and segmented images and clinical metadata (sex, age, localization).
\end{itemize}

Table~\ref{tab:ablation_study} presents the test accuracy of these configurations on the HAM10000 and ISIC 2019 datasets.

\begin{table}[t]
\centering
\caption{Ablation Study Results on HAM10000 and ISIC 2019 Datasets}
\label{tab:ablation_study}
\renewcommand{\arraystretch}{1.2}
\setlength{\tabcolsep}{10pt}
\begin{tabular}{|l|c|c|}
\hline
\textbf{Model} & \textbf{HAM10000 (\%)} & \textbf{ISIC 2019 (\%)} \\
\hline
DenseNet201 (Original) & 88.88 & 84.17 \\
DenseNet201 (Segmented) & 84.74 & 79.23 \\
Dual-Network (No Metadata) & 91.23 & 87.14 \\
\textbf{Dual-Network (With Metadata)} & \textbf{93.47} & \textbf{88.56} \\
\hline
\end{tabular}
\end{table}

As shown in Table~\ref{tab:ablation_study}, the inclusion of both original and segmented images significantly improves performance. Furthermore, integrating clinical metadata (sex, age, and localization) enhances accuracy, indicating that metadata provides complementary information that boosts the model’s performance.

\begin{figure}[h!]
    \centering
    \begin{minipage}{0.48\textwidth}
        \centering
        \includegraphics[width=\textwidth]{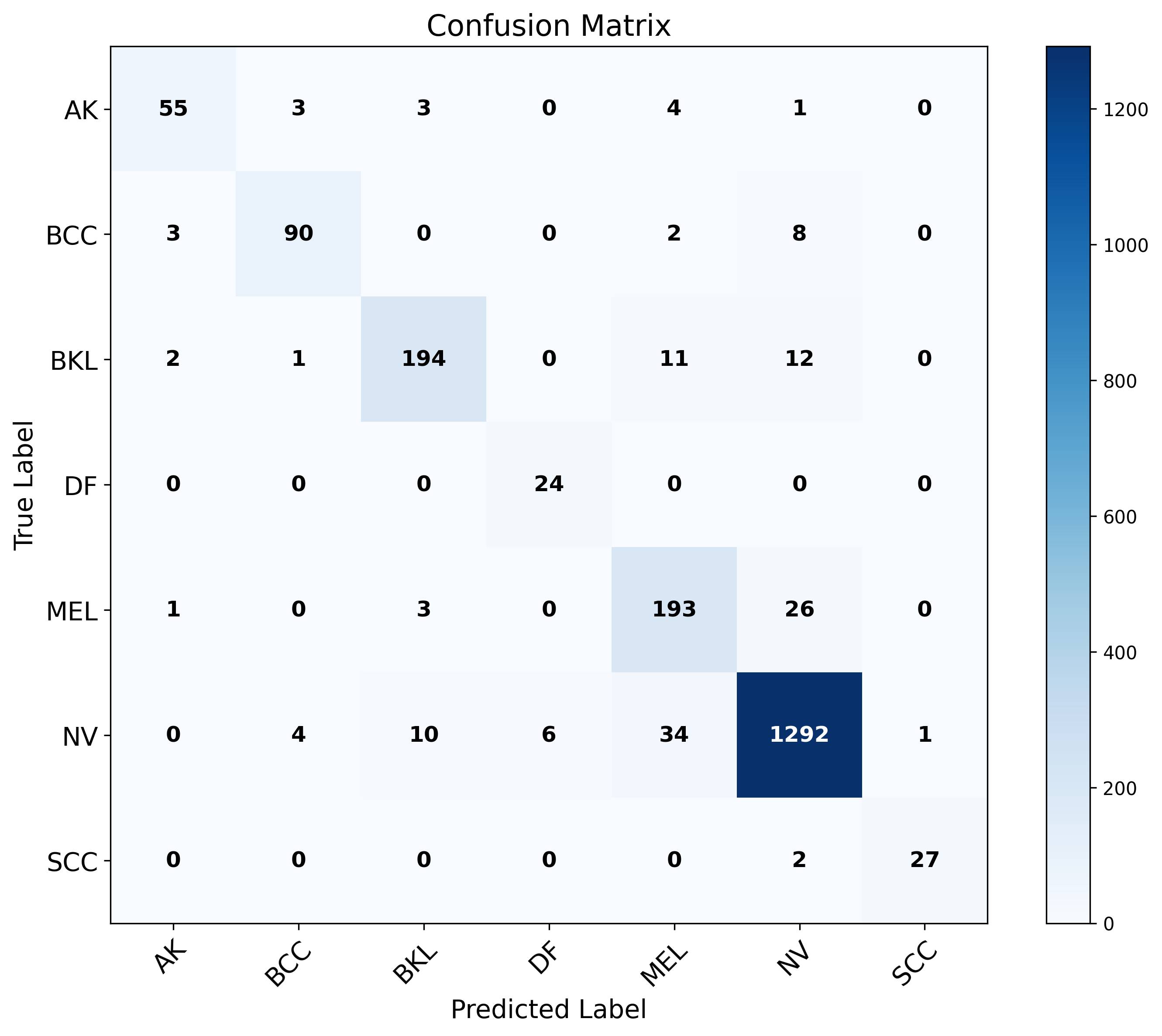}
    \end{minipage}
    \hfill
    \begin{minipage}{0.48\textwidth}
        \centering
        \includegraphics[width=\textwidth]{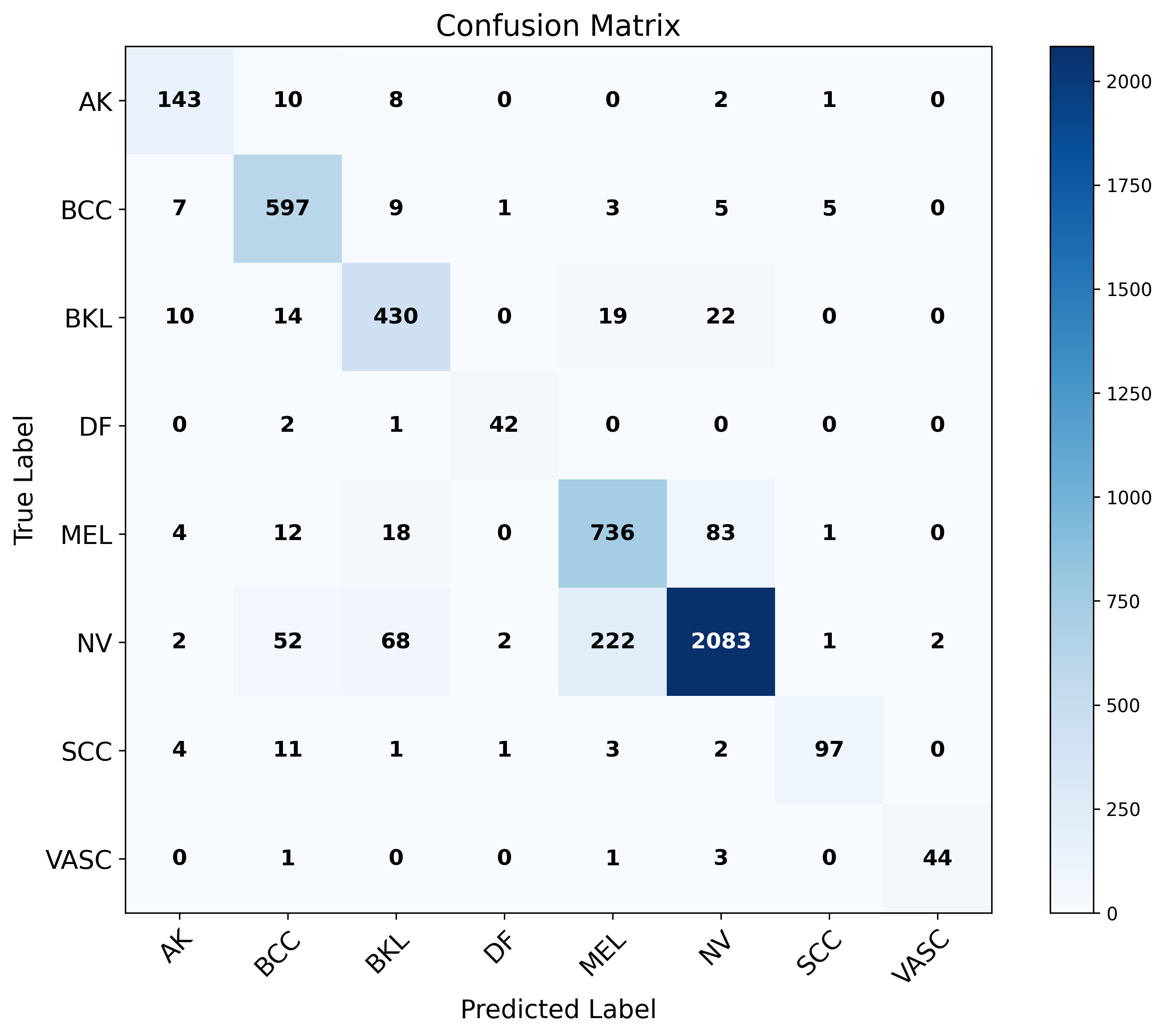}
    \end{minipage}
    \caption{Confusion Matrix on HAM10000 (left) and ISIC 2019 (right) Test Set}
    \label{fig:confusion-matrix}
\end{figure}

Fig.~\ref{fig:confusion-matrix} shows the confusion matrices for the HAM10000 and ISIC 2019 datasets, respectively. These matrices provide further insight into the model's class-wise performance and indicate balanced sensitivity across most lesion classes.

\subsection{Model Explainability and Reliability with Grad-CAM}
Beyond quantitative metrics, it is crucial to ensure that a model's predictions are based on clinically relevant features rather than spurious correlations from the background. To validate this, we employed Gradient-weighted Class Activation Mapping (Grad-CAM) to visualize the regions of the input image that were most influential in the model's classification decision.

\begin{figure}[h!]
    \centering
    \includegraphics[width=\textwidth]{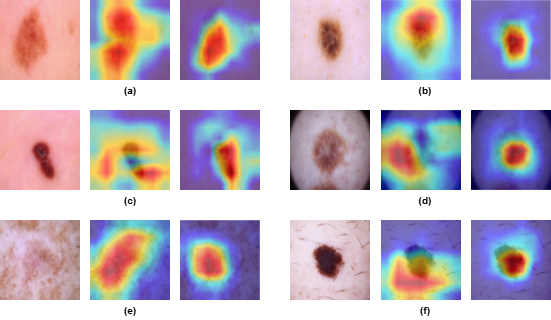}
    \caption{Grad-CAM visualizations for four different examples. Each set (a-f) shows (left to right): the Original Image, the heatmap from a baseline DenseNet201, and the heatmap from our Proposed Model. Our model's activations are consistently and precisely focused on the lesion area, unlike the baseline's diffuse or misplaced focus, demonstrating superior explainability and reliability.}
    \label{fig:gradcam_comparison}
\end{figure}

We compare the Grad-CAM heatmaps generated by a standard DenseNet201 model with those from our proposed dual-input model in Fig.~\ref{fig:gradcam_comparison}. The results clearly demonstrate the enhanced explainability and reliability of our approach. The heatmaps from the baseline DenseNet201 model are often diffuse, with activation spread across both the lesion and irrelevant background areas. In some cases, the baseline model focuses on image artifacts or edges rather than the core of the lesion.

In contrast, the heatmaps generated by our proposed model show a remarkably focused and precise activation that is tightly concentrated on the lesion area. This is a direct result of our architecture: by providing the segmented lesion as a second input and using a cross-attention mechanism, we explicitly guide the model to prioritize features within the lesion boundaries. This focused attention confirms that our model's superior accuracy is not coincidental but is derived from learning the correct pathological features. This high degree of explainability builds confidence in the model's decisions, making it a more reliable tool for potential clinical applications.

\section{Conclusion}

This paper presents a novel framework for automated skin cancer classification designed to address the critical needs for both high accuracy and model interpretability. The proposed approach integrates a Deep-UNet for precise lesion segmentation with a dual-input classification model that effectively fuses global, lesion-specific, and clinical information. By utilizing two parallel DenseNet201 encoders for original and segmented images, a Multi-Head Cross-Attention (MHCA) module facilitates a rich interaction between them. This architecture successfully guides the model to focus on clinically relevant pathological features within the lesion.

To further enhance diagnostic accuracy, clinical metadata is seamlessly integrated using a transformer-based encoder, allowing the model to contextualize visual features with patient-specific attributes. The final fused representation, capturing a holistic view of both visual and clinical data, is processed by a multi-layer classifier to generate the final prediction. This synergistic combination of segmentation-guided feature extraction and clinical data integration is the cornerstone of the model's improved performance.

Experiments on the challenging HAM10000 and ISIC 2019 datasets validate our approach. The proposed model not only outperforms several state-of-the-art architectures in accuracy and AUC but also demonstrates enhanced reliability. Crucially, through Gradient-weighted Class Activation Mapping (Grad-CAM), we have provided visual evidence that our model's decision-making process is concentrated on the lesion area, unlike baseline models that often rely on spurious background features. This confirmation of the model's focus addresses the "black box" problem, significantly boosting its trustworthiness and potential for real-world dermatological applications.

Future work will aim to enhance the model's generalizability and prepare it for clinical translation. Potential directions include incorporating a wider range of clinical and patient history data, such as genetic markers, ethnicity, and UV exposure history. Furthermore, we plan to conduct prospective validation on external, unseen datasets and explore other advanced explainability methods to provide even deeper insights into the model's diagnostic reasoning.

\bibliographystyle{elsarticle-num-names}  
\bibliography{references}  
\end{document}